\documentclass[10pt,twocolumn,letterpaper]{article}

\usepackage[pagenumbers]{cvpr} % To force page numbers, e.g. for an arXiv version
\usepackage{hyperref}
\usepackage[dvipsnames]{xcolor}

\hypersetup{
    colorlinks = true,
    linkbordercolor = {white},
    citecolor=blue,
    urlcolor=red
}

\definecolor{cvprblue}{rgb}{0.21,0.49,0.74}
\usepackage{amsmath}
\usepackage{amssymb}
\usepackage{mathtools}

\def\paperID{*****} % *** Enter the Paper ID here
\def\confName{CVPR}
\def\confYear{2024}

\title{
StereoCrafter: Diffusion-based Generation of Long and High-fidelity Stereoscopic 3D from Monocular Videos}
\author{
Sijie Zhao$^*$ \quad
Wenbo Hu$^*$ \quad
Xiaodong Cun$^*$ \quad
Yong Zhang$^\dag$ \quad
Xiaoyu Li$^\dag$ \quad \\
Zhe Kong  \quad
Xiangjun Gao  \quad
Muyao Niu  \quad
Ying Shan \quad
\\\\
Tencent AI Lab \quad
ARC Lab, Tencent PCG
\\\\
{{Project page: {\url{http://stereocrafter.github.io}}}} \\
}

\begin{document}

% \maketitle
\twocolumn[{
\maketitle
\begin{center}
    \captionsetup{type=figure}
    \includegraphics[width=1.\textwidth]{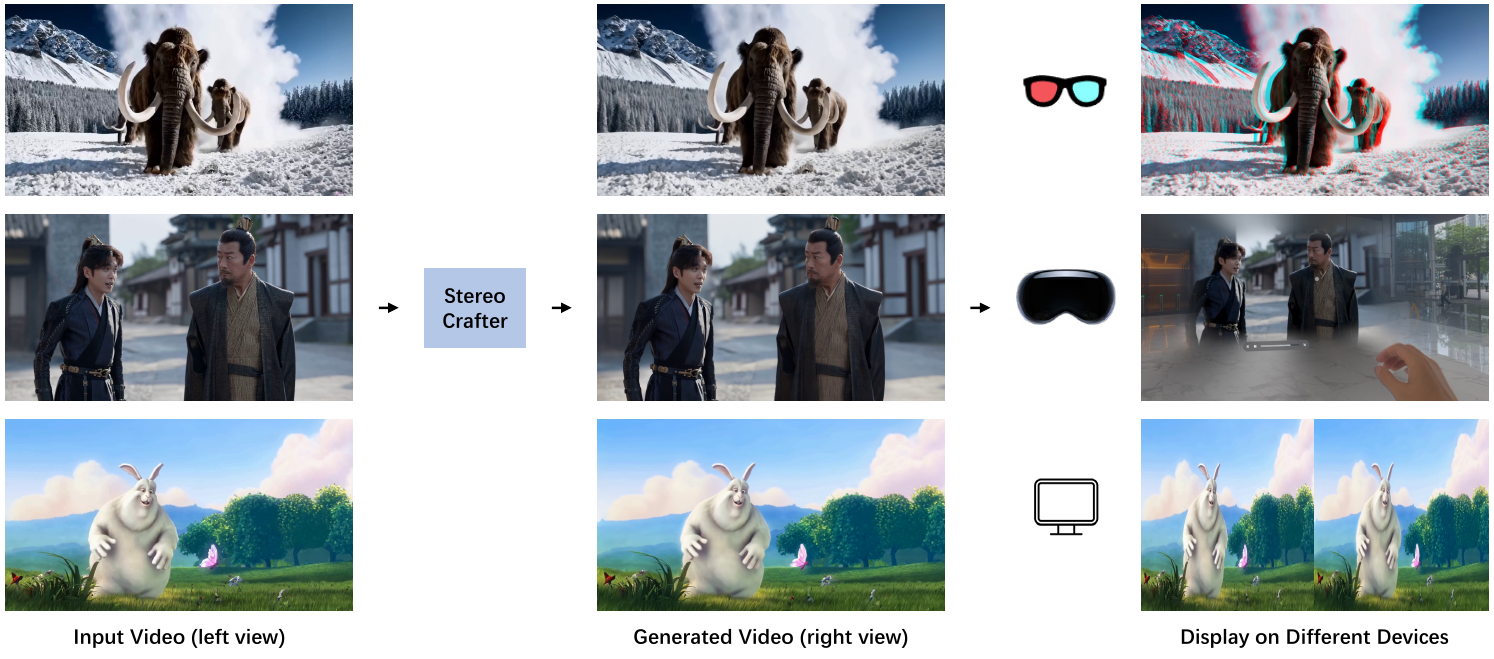}
    \vspace{-2em}
    \captionof{figure}{We propose a framework to convert any 2D videos to immersive stereoscopic 3D ones that can be viewed on different display devices, like 3D Glasses, Apple Vision Pro and 3D Display. It can be applied to various video sources, such as movies, vlogs, 3D cartoons, and AIGC videos. We hope this approach can be applied to revolutionize the way we experience digital media in the future.}
    \label{fig:teaser}
\end{center}
}]

{
  \renewcommand{\thefootnote}%
    {\fnsymbol{footnote}}
  \footnotetext[1]{~Equal contribution.} \footnotetext[2]{~Corresponding author.}
}

\begin{abstract}
This paper presents a novel framework for converting 2D videos to immersive stereoscopic 3D, addressing the growing demand for 3D content in immersive experience. Leveraging foundation models as priors, our approach overcomes the limitations of traditional methods and boosts the performance to ensure the high-fidelity generation required by the display devices. The proposed system consists of two main steps: depth-based video splatting for warping and extracting occlusion mask, and stereo video inpainting. We utilize pre-trained stable video diffusion as the backbone and introduce a fine-tuning protocol for the stereo video inpainting task. To handle input video with varying lengths and resolutions, we explore auto-regressive strategies and tiled processing. Finally, a sophisticated data processing pipeline has been developed to reconstruct a large-scale and high-quality dataset to support our training. Our framework demonstrates significant improvements in 2D-to-3D video conversion, offering a practical solution for creating immersive content for 3D devices like Apple Vision Pro and 3D displays. In summary, this work contributes to the field by presenting an effective method for generating high-quality stereoscopic videos from monocular input, potentially transforming how we experience digital media.
\end{abstract}

\vspace{-0.15cm}
\section{Introduction}
\label{sec:intro}
Pursuing a more immersive experience of digital content is gaining growing popularity due to the captivating and enjoyable psychological feeling of spatial presence. In contrast to traditional 2D digital content, immersive 3D content is becoming the next frontier, owing to the advancements in virtual reality (VR) and augmented reality (AR) technology in both hardware, such as the release of Apple Vision Pro, and software, including vision foundation models. However, a substantial volume of digital media like video clips, teleplays and movies on the internet are monocular which lack vividness for displaying in 3D, contrasted with the limited availability of 3D video content. Consequently, the conversion of these 2D videos into immersive 3D videos has been highly demanded.

The human visual system utilizes parallax between the left and right eye images to gain depth perception such that scenes are perceived in three dimensions rather than on a two-dimensional plane. Consequently, 3D videos are typically represented in a stereoscopic format. 2D-to-3D video conversion methods~\cite{zhang20113d, konrad2013learning} in the early stage usually consist of two main steps: the extraction of depth information from the input view and the rendering of a novel view based on the depth (Depth Image Based Rendering) to form a stereo pair. Deep3D~\cite{xie2016deep3d} suggests directly regressing the right view using a pixel-wise loss, by predicting a probabilistic disparity-like map as an intermediary output. Nonetheless, owing to the limited training data and model capacity of convolutional neural networks, these methods tend to produce blurry results with limited generalization ability to real-world videos, which is far from the practice usage.

Recently, the emergence of 3D representations such as Neural Radiance Fields~\cite{mildenhall2021nerf} (NeRF) and 3D Gaussian Splatting~\cite{kerbl20233d} (3DGS) have significantly transformed the field of novel view synthesis due to their high-quality results and simple reconstruction processes. Hence, an alternative approach for 2D-to-3D video conversion involves reconstructing the dynamic 3D scene from the input video and generating stereoscopic videos via novel view synthesis. However, these methods~\cite{park2021nerfies, park2021hypernerf, li2021neural, li2023dynibar, liu2023robust, lee2023fast} require estimating the camera pose of each frame from monocular videos, which heavily rely on the static parts in the video for calibration. For videos exhibiting large camera motion, sizable dynamic objects, or visual effects such as fog or fire, calibrating the cameras and reconstructing the scenes becomes a challenging task for these methods. In addition, these methods usually handle the occluded regions by blending the information from neighboring frames and cannot address the occlusion that does not appear in the neighboring frames. Consequently, we believe that dynamic 3D reconstruction methods are not a practical and effective solution for producing stereoscopic videos.

Meanwhile, trained from large-scale data, foundation models~\cite{radford2021learning, caron2021emerging, rombach2022high, touvron2023llama} have emerged and garnered lots of attention due to their strong zero-shot performance in various downstream tasks. Benefiting from these basic models, recent works for monocular depth estimation from a single image~\cite{ke2024repurposing, yang2024depth, yin2023metric3d, piccinelli2024unidepth} have demonstrated remarkable results with substantial improvements compared with traditional methods. On the other hand, the utilization of basic video diffusion models has boosted the performance of related tasks such as video generation~\cite{blattmann2023stable,niu2024mofa}, video editing~\cite{qi2023fatezero, ma2023follow} and video inpainting~\cite{zhou2023propainter,liu2023coordfill}. These developments inspire us to rethink the problem of 2D-to-3D video generation, which does not have a practical solution due to the limited performance in depth estimation and inpainting with traditional methods.

Therefore, by leveraging foundation models as model priors, we present a framework for converting 2D videos of various types to stereoscopic 3D which could be immersively experienced with devices like Apple Vision Pro and 3D displays as shown in Fig.~\ref{fig:teaser}. We introduce a practical solution for 2D-to-3D video conversion and achieve usable quality for the industry. Our system consists of two main steps: depth-based video splatting and stereo video inpainting. We first employ a depth estimation method to give us depth maps of the input video. Utilizing this depth map, we warp the input video from the left view to the right view via a depth-based video splatting method, which concurrently produces an occlusion mask. Subsequently, based on the warped video and its corresponding occlusion mask, we generate the final right-view video using our stereo video inpainting method. 

To generalize our framework to input videos with various types, we first employ pre-trained stable video diffusion~\cite{blattmann2023stable} as the backbone of our network. Leveraging this diffusion prior trained on a large-scale dataset, video quality and consistency of the results could be greatly ensured. Subsequently, we propose a fine-tuning protocol to adapt the model for the stereo video inpainting task, which requires data comprising occluded videos, occlusion masks and complemented videos. To reconstruct this dataset, we present a data processing pipeline that utilizes our video splatting approach based on collected stereo videos. Finally, to adapt the model to the input videos with varying lengths and resolutions, an auto-regressive strategy and tiled processing are explored in our work.

Our major contributions can be summarized as follows:
\begin{itemize}
    \item We develop a framework for converting 2D videos to stereoscopic 3D with immersive experience leveraging diffusion model priors and our reconstructed dataset.
    \item We design a data processing pipeline to facilitate the training of our approach with high-quality data.
    \item We introduce a depth-based video splatting that could produce accurate warped videos and occlusion masks in parallel for each pixel, which has a fast processing speed running on modern GPUs.
    \item We propose a stereo video inpainting network with an auto-regressive strategy and tiled processing to handle input video with different lengths and resolutions.
\end{itemize}

\section{Related Work}
\label{sec:related-work}
%--------------------------------------

\paragraph{2D-to-3D Video Conversion.}
2D-to-3D video conversion has been an active area of research in computer vision and graphics since the popularity of head-mounted displays for 3D content and 3D movie production. Early approaches~\cite{zhang20113d, konrad2013learning} for 2D-to-3D video conversion relied on depth estimation and image-based rendering techniques. Specifically, Deep3D~\cite{xie2016deep3d} introduces an end-to-end network to directly generate the right view from the left view.~\citet{lang2010nonlinear} propose a method to retarget stereoscopic 3D video automatically to a novel disparity range. Other approaches leverage deep learning methods for video depth estimation~\cite{zhang2021consistent, kopf2021robust} and utilize the depth for novel view synthesis. Recently, some methods have explored the use of diffusion models for various tasks including stereo generation~\cite{wang2024stereodiffusion, dai2024svg}. Despite these advancements, generating high-quality, consistent stereoscopic videos from monocular input remains challenging, especially for scenes with complex motion or occlusions.

\paragraph{Dynamic View Synthesis from Monocular Videos.}
Recent advances in 3D reconstruction, such as NeRF~\cite{mildenhall2021nerf} and 3D Gaussian Splatting~\cite{kerbl20233d}, have greatly improved the quality of novel view synthesis, which also facilitate the view synthesis for a dynamic scene from monocular videos~\cite{park2021nerfies, li2021neural, gao2021dynamic, liu2023robust, lee2023casual-fvs, li2023dynibar, wang2024gflow, tretschk2021non}. By reconstructing dynamic scenes, these methods can synthesize space-time results, which can also be utilized for creating stereoscopic videos. However, relying on camera poses as input or jointly optimizing camera poses within the method makes it challenging to handle complex dynamic scenarios where camera poses are hard to optimize.

\paragraph{Video Diffusion Models.} The video generation methods have rapid development in recent studies due to the stronger abilities of the diffusion model~\cite{ho2020denoising}. Early works~\cite{ho2022video, he2022latent} directly train the multi-scale video diffusion models from the video data. Thanks to the large-scale pre-trained text-to-image model, \ie, Stable Diffusion~\cite{podell2023sdxl, rombach2022high}, adding temporal layers to the text-to-image models for text-to-video generation are also popular~\cite{hu2022make, ho2022imagenvideo, videocrafter1, chen2024videocrafter2, zhou2022magicvideo, wang2023lavie}. More recently, 3D-VAE~\cite{zhao2024cvvae} based video diffusion model, \ie, Sora~\cite{sora} show more advanced results on this topic. These text-to-video diffusion models provide a strong visual backbone for other conditional generation tasks. \eg, the image-to-video generation~\cite{blattmann2023stable,niu2024mofa}, video-editing~\cite{qi2023fatezero, ma2023follow}, video-to-video translation and enhancement~\cite{xing2023make, yang2024noise}, frame-interpolation~\cite{yang2024zerosmooth}.  
In this paper, we also utilize the pre-trained knowledge from the video diffusion model for our stereo video generation task.

\section{Methodology}
\label{sec:method}
%--------------------------------------

\begin{figure*}[ht!]
\centering
\includegraphics[width=1.0\textwidth]{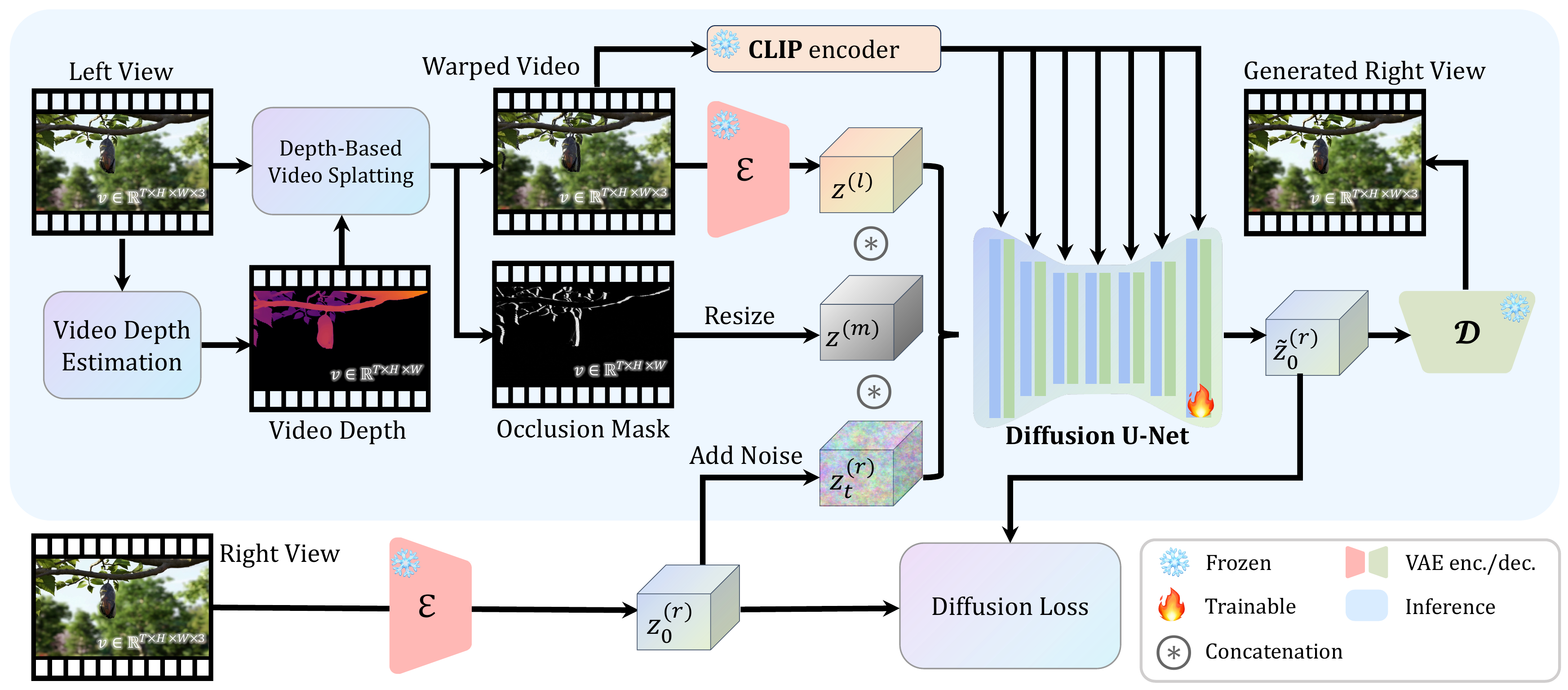}
\caption{Overall framework of StereoCrafter, which contains two main stages. In the first stage, the video depth is estimated from the monocular video and we obtain the warped video and its occlusion mask through depth-based video splatting with the left video and the video depth as input. Then, we train a stereo video inpainting model to fill in the hole region of the warped video according to the occlusion mask to synthesize the right video. }
\label{fig:framework}
\end{figure*}

\subsection{Overview}
We propose a stereo video generation framework that converts a monocular video into a stereo video for an immersive experience, which can be viewed in VR/AR devices or 3D display. As shown in Fig.~\ref{fig:framework}, the framework consists of two main stages: depth-based video splatting and stereo video inpainting. We first determine the depth of the input monocular video by utilizing a video depth estimation model and perform depth-based video splatting to warp the input video from the left view to the right view, simultaneously obtaining the corresponding occlusion mask. Then, we train a diffusion model for stereo video inpainting to fill the holes of the warped video based on the occlusion mask, resulting in the final right view. The input left and completed right videos can be viewed on stereo display devices like Vision Pro. Finally, we present our data processing pipeline designed for constructing the training dataset, which significantly contributes to our success.

\subsection{Depth-based Video Splatting}
\label{subsec:warping}
%--------------------------------------
We utilize disparity maps to synthesize right-view videos from the left-view input, necessitating a depth estimation method to predict the depth of the input video. Numerous depth estimation methods have been proposed in the past~\cite{eigen2014depth, fu2018deep, lee2019big, aich2021bidirectional, li2023depthformer, yang2021transformer, patil2022p3depth}, and significant progress has been made recently in this field~\cite{ranftl2020towards, bhat2023zoedepth, yin2023metric3d, piccinelli2024unidepth, depthanything, depth_anything_v2, marigold, fu2024geowizard} benefiting from the utilization of model priors, such as stable diffusion and DINO. Consequently, we adopt the state-of-the-art depth estimation method DepthCrafter~\cite{hu2024-DepthCrafter} or Depth Anything V2~\cite{depth_anything_v2} to obtain detailed video depth for the input video. DepthCrafter~\cite{hu2024-DepthCrafter} is capable of producing more temporally consistent video depth results than Depth Anything V2~\cite{depth_anything_v2}, making it a better fit for our task.

%The video depth estimation model is trained with the pre-trained weights of SVD~\cite{blattmann2023stable} as initialization, which exploits the video priors instead of training from scratch. The monocular video is concatenated with the noise as the input of the UNet of SVD, while the output is the video depth. To construct the training data, we gathered stereo videos from the internet and split them into individual shots. Next, we apply the video stereo matching method~\cite{jing2024match} to extract pseudo video depth for training purposes. The training data comprises 180K clips, each with extracted video depth.
 
\begin{figure}
\centering
\includegraphics[width=1.0\linewidth]{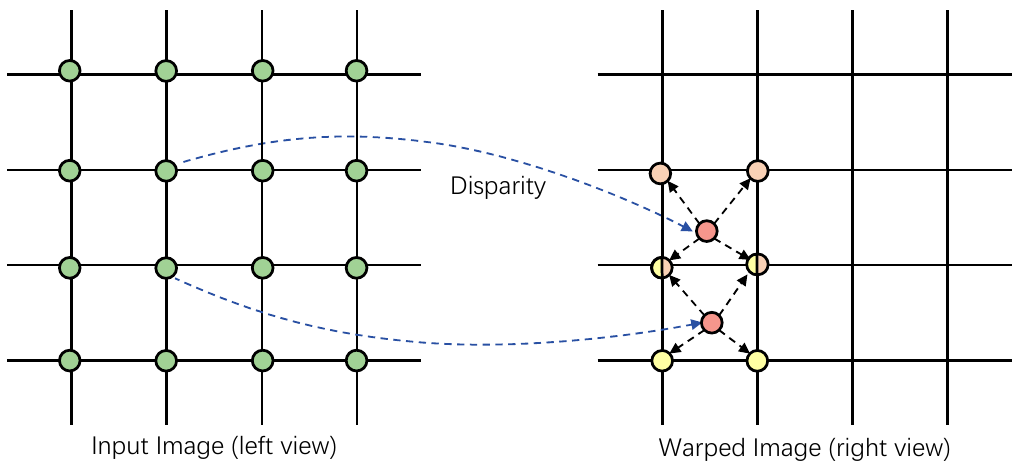}
\caption{Illustration of our depth-based forward splatting. The image on the right is created by splatting the input pixels according to the disparity. And we use a depth-aware method to resolve any ambiguity when multiple pixels are splatted to the same pixel in the right view.}
\label{fig:splatting}
\end{figure}

%--------------------------------------
After estimating the depth of the input video, our goal is to warp the input left video to the right view, using the disparity calculated from the depth. The disparity map indicates the target position for each pixel in the left image, as it transitions from the left to the right image. It is important to note that disparity is a forward mapping and common resampling techniques such as backward warping and interpolation cannot be applied to it. Consequently, we propose a forward splatting method capable of mapping each source pixel to the target image based on its disparity as shown in Fig.~\ref{fig:splatting}.

During forward splatting, we map each pixel in the source image to its target position and splat it onto the four nearest pixels in the target image's grid based on its distance to the target position. However, multiple pixels from the source image may be mapped to the same pixel in the target image, creating ambiguity that requires a proper solution as illustrated in Fig.~\ref{fig:splatting}. To address this issue, we calculate the weights of the splatted pixel based on the disparity value of its source pixel. We then accumulate all the splatted pixels corresponding to the same target pixel and blend them according to their weights to obtain the final color of the target pixel. It is important to note that a large disparity signifies a smaller depth of the pixel, indicating that it is closer than other pixels. As a result, our method assigns a larger blending weight to such pixels. Therefore, we calculate the weight according to the formula $w = \sqrt{2}^{disp}$. For target pixels without any splatted pixels, we mark them as occlusion pixels and calculate the occlusion mask for the subsequent inpainting process. We have implemented a parallel version of our splatting approach on the GPU, which can run in real-time on modern GPUs.

\subsection{Stereo Video Inpainting}
\label{subsec:inpainting}

\begin{figure*}
\centering
\includegraphics[width=1.0\linewidth]{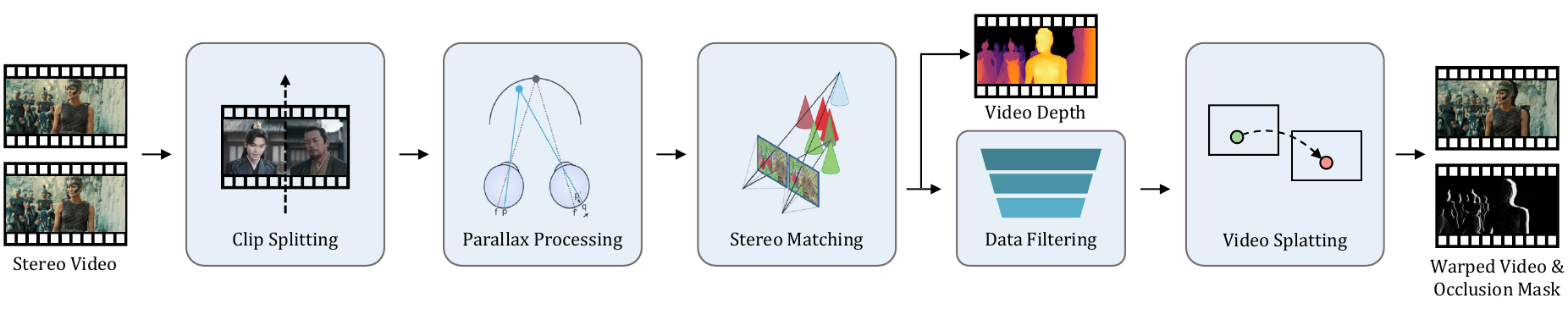}
\caption{The pipeline of our approach for constructing the training dataset. After curating a large number of stereo videos, we generate the video depth/disparity, warped left video, and occlusion mask for each data sample, while using the right video as the ground truth.}
\label{fig:dataset}
\end{figure*}

\begin{figure}
\centering
\includegraphics[width=1.0\linewidth]{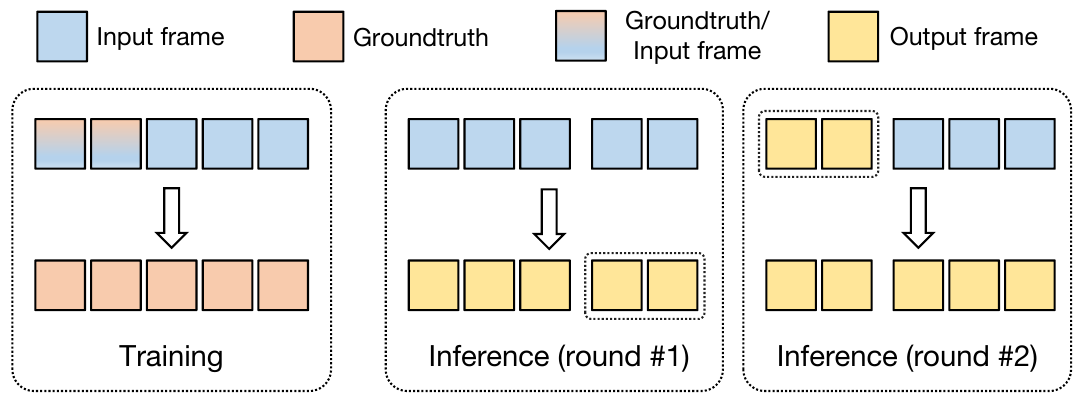}
% \caption{During the training process, we randomly replace part of the input frames with ground truth, which allows us to maintain the consistency of the inpainting area in multi-round inference. As a result, we can achieve inpainting for videos of arbitrary length.}
\caption{Illustration of our approach for handling videos of arbitrary length.}
\label{fig:inpainting_longvideo}
\end{figure}

\begin{figure}
\centering
\includegraphics[width=1.0\linewidth]{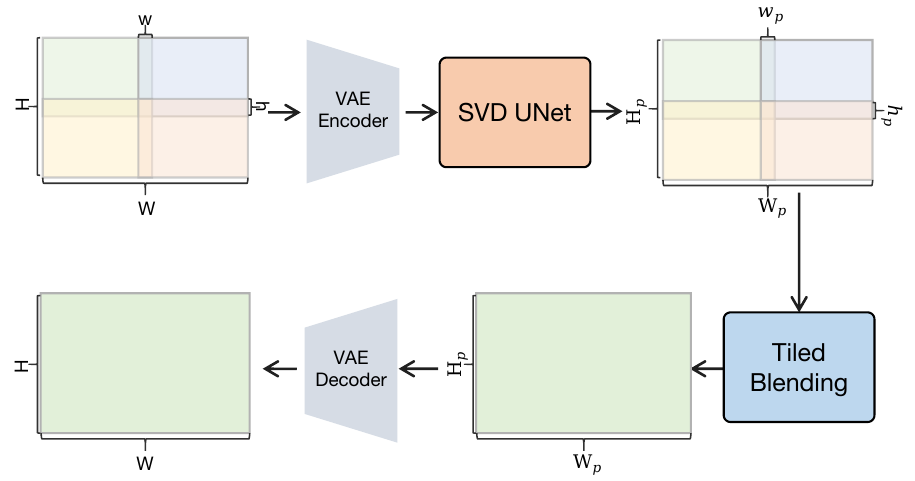}
\caption{Illustration of our approach for handling high resolution input.}
\label{fig:inpainting_superresolution}
\end{figure}

%--------------------------------------
Given the warped video and corresponding occlusion mask, we introduce a stereo video inpainting method to address the occluded pixels and synthesize the output right-view video. As shown in Fig.~\ref{fig:framework}, we extend the Stable Video Diffusion (SVD) for stereo video inpainting, which includes: (1) changing the condition of SVD from image to warped frames; (2) adding an extra channel in the input layer of Unet (i.e., increasing from 8 to 9) to input the occlusion mask, and we set the parameters corresponding to this channel in the first convolutional layer to 0 for zero initialization.

Based on our video inpainting model, we propose the following methods to achieve stereo video inpainting of arbitrary length and resolution while keeping the consistency of the generated results. (1) \textbf{Auto-regressive modeling}. A common video clip may have hundreds of frames, while the SVD released version can only generate 25 frames. Therefore, multiple inferences are needed when the input video is longer. However, simply splitting the video along the time dimension and processing it independently will result in inconsistencies in the inpainting area between adjacent blocks. Therefore, we propose auto-regressive modeling to deal with videos of arbitrary length. As shown in Fig.~\ref{fig:inpainting_longvideo}, during the training process, we randomly sample a value $n$ from $0$ to $N$ and replace the first $n$ input frames with ground truth. Then we use the standard diffusion loss for training. During the inference process, we concatenate the last $m$ frames generated in the previous round with the subsequent input frames as the input for the next round. Therefore, the model can generate more temporally consistent contents by combining the inpainting results from the previous round. (2) \textbf{Tiled diffusion}. Video diffusion models require a large amount of memory during the inference process, making it difficult to process high-resolution videos in limited memory. Therefore, we propose tiled diffusion for high-resolution video inpainting. As shown in Fig.~\ref{fig:inpainting_superresolution}, We first divide the high-resolution video into blocks along the spatial dimension, then use the video diffusion model to independently infer each block, and then blend the overlap areas of adjacent blocks in the latent space. Taking horizontal blending as an example, $blended = \mathbf{w} * left + (1 - \mathbf{w}) * right$, where $\mathbf{w}$ increases from 0 to 1 in the horizontal direction. Afterwards, we use VAE for decoding to obtain the inpainting results at the target resolution. With this tiled diffusion processing, we could break through the memory limitation to process videos at high resolution.

\subsection{Dataset Construction}
\label{subsec:dataset}

% \twocolumn[{
% \begin{center}
%     \captionsetup{type=figure}
%     \includegraphics[width=1.0\textwidth]{SVG/figs/stereo_comparison_v1.pdf}
%     \captionof{figure}{Qualitative comparison results of our approach with different 2D-to-3D conversion methods. Our approach could synthesize high-quality results while maintaining consistency with the input left view from the stereo matching results, which is important for display in 3D devices.}
%     \label{fig:stereo_comp}
% \end{center}
% }]

\begin{figure*}
\centering
\includegraphics[width=1.0\linewidth]{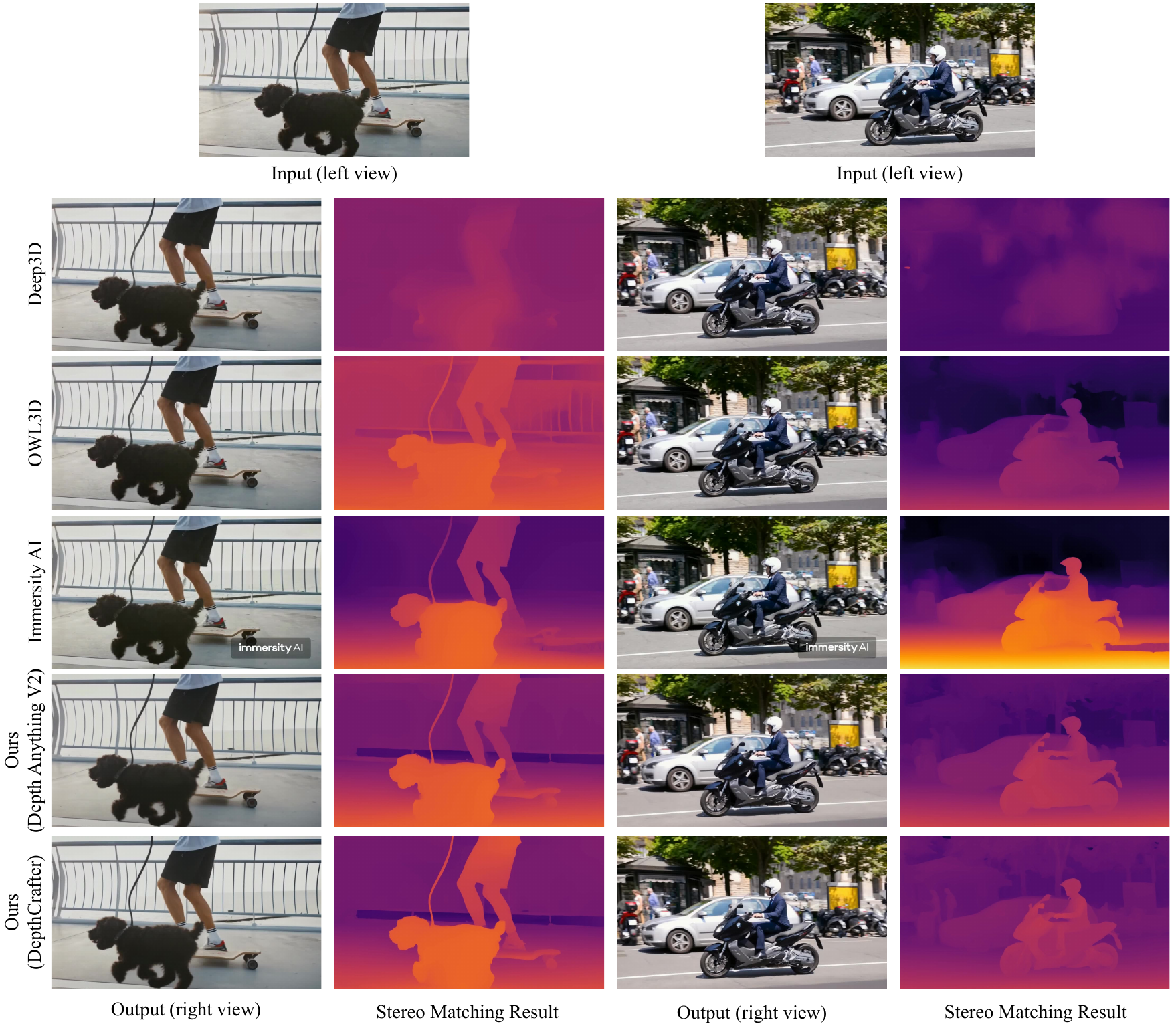}
\caption{Qualitative comparison results of our approach with different 2D-to-3D conversion methods. Our approach could synthesize high-quality results using different depth estimation methods like Depth Anything V2~\cite{depth_anything_v2} or DepthCrafter~\cite{hu2024-DepthCrafter}, while maintaining consistency with the input left view as shown by the matching results. }
\label{fig:stereo_comp}
\end{figure*}

To boost the performance of our stereo video generation methods, an appropriate dataset is required for training, as the quality of the dataset significantly influences the fidelity of the results produced by diffusion models. However, there is no existing dataset that we could use directly, which should include a warped left video and an occlusion mask as inputs, a completed right video as the ground truth for each data sample. 

As illustrated in~Fig.~\ref{fig:dataset}, the data pipeline begins with curating a diverse collection of stereo videos spanning various categories. We employed the PySceneDetect tool to perform shot detection within each stereo video, facilitating automated segmentation into individual clips. Subsequently, the stereo matching method outlined in~\cite{jing2024match} was applied to predict disparity maps between the left and right views of each clip, enabling accurate reconstruction of stereo content.

However, stereo videos in the wild may exhibit varying disparity ranges based on the distinct definitions of the zero disparity plane in the scene. Directly feeding these videos to video stereo matching methods can result in inaccurate results, as the learning-based methods are typically trained on a specific dataset within a particular disparity range. To address this issue, we employ a parallax processing step before stereo matching. Specifically, we shit the right view video to the left by a certain amount and crop the left view video accordingly until the disparity for all pixels is negative and the maximum disparity is nearly zero. After this processing, we align the disparity distribution of the stereo video to the training data of the video stereo matching methods, thereby yielding more accurate matching results.

After obtaining the disparity map, we warp the left view video to the right view using our depth-based video splatting method, which will output the warped video and its corresponding occlusion mask. In addition, to exclude the data with large stereo matching errors, we filter the data by calculating the PSNR between the right view video and the warped left view video and only retain samples with PSNR greater than 25dB. Through the above processing steps, we have collected approximately 180k training sequence samples with about 25 million frames.

\section{Experiments}
\label{sec:exp}

% \twocolumn[{
% \begin{center}
%     \captionsetup{type=figure}
%     \includegraphics[width=1.0\textwidth]{SVG/figs/inpainting_comparison_v1.pdf}
%     \captionof{figure}{Quantitative comparison results of different video inpainting models.}
%     \label{fig:inpainting_comp}
% \end{center}
% }]

\begin{figure*}
\centering
\includegraphics[width=1.0\textwidth]{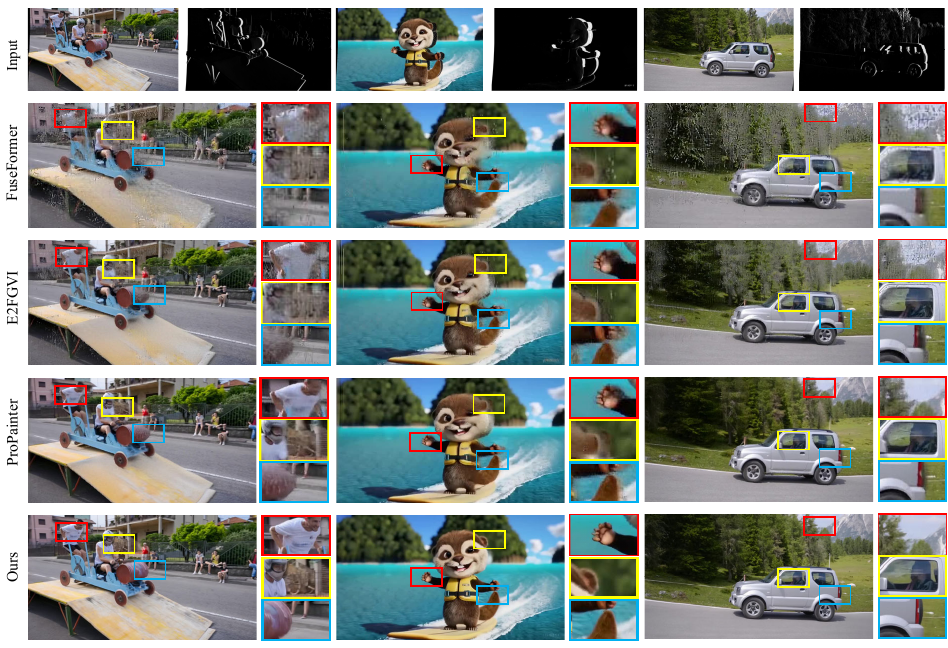}
\caption{Qualtitative comparison results of different video inpainting models. Our method is capable of producing sharper results in the occluded areas.}
\label{fig:inpainting_comp}
\end{figure*}

\subsection{Implementation Details}
%--------------------------------------
\paragraph{Datasets.}
We curate a large number of stereo videos as the data sources, which are cropped from the long video into different clips and decoded into left view videos $V_{left}$ and right view videos $V_{right}$. To get pseudo ground truth depth, we calculate the disparity map between the left view video and right view video using the video stereo matching method~\cite{jing2024match}, i.e., $D_{left} = Matching(V_{left}, V_{right})$. Subsequently, based on this disparity map, we perform forward splatting on the left view video according to the method in Sec.~\ref{subsec:warping}, obtaining the warped video and occlusion mask, i.e., $V_{warped}, M_{occlusion} = Splatting(V_{left}, D_{left})$. Ultimately, the training data pairs required for stereo video inpainting are formed by $(V_{warped}, M_{occlusion}, V_{right})$.

\paragraph{Training Details.}
We train our stereo inpainting model based on the aforementioned dataset. We initialize the model with pre-trained weights from SVD and only finetune the spatial layers in the U-Net. We sample the training data from the dataset with a resolution of $25\times 576\times 1024$ and a frame stride ranging from 1 to 6. We used a constant learning rate of 1e-5 with the AdamW~\cite{loshchilov2017decoupled} optimizer. The training is conducted on 8 A100 GPUs with a batch size of 1 per GPU and 26K iterations. For training efficiency, we employ deepspeed stage 2~\cite{rajbhandari2020zero}, gradient checkpointing~\cite{chen2016training} techniques, and train with float16 precision.

\subsection{Comparison to 2D-to-3D Video Conversion}
\label{subsec:stereo}
Firstly, We compare our framework with traditional 2D-to-3D video conversion methods Deep3D~\cite{xie2016deep3d} and some 2D-to-3D conversion software Owl3D~\cite{owl3d} and Immersity AI~\cite{ImmersityAI}. In particular, Deep3D~\cite{xie2016deep3d} proposes a fully automatic 2D-to-3D conversion approach that is trained end-to-end to directly generate the right view from the left view using convolutional neural networks. Owl3D~\cite{owl3d} is an AI-powered 2D to 3D conversion software and Immersity AI is a platform converting images or videos into 3D. For Owl3D and Immersity AI, we upload the input left view videos to their platform and generate the right view video for comparison. The qualitative comparison results are shown in Fig.\ref{fig:stereo_comp}. In addition to showing the right view results, we also employ a video stereo matching approach~\cite{jing2024match} to estimate the disparity between input left view video and output right view video to verify its spatial consistency. As shown in Fig.\ref{fig:stereo_comp}, Deep3D~\cite{xie2016deep3d} could generate overall promising right view results, but is not spatially consistent with the input video according to the stereo matching results. On the other hand, Owl3D and Immersity AI could generate more consistent results, but some artifacts appear in the images, such as the handrail in the first example. In the end, our method could synthesize high-quality image results while keeping consistency with the left view images from the stereo matching results using different depth estimation methods. With more temporally consistent video depth predicted by DepthCrafter, our method could achieve even better results.

\subsection{Comparison to Video Inpainting}
\label{subsec:inpainting}

We show the qualitative results of our method on the stereo video inpainting and compare it with previous video inpainting models, including FuseFormer~\cite{liu2021fuseformer}, E2FGVI~\cite{liCvpr22vInpainting} and ProPainter~\cite{zhou2023propainter}. As shown in Fig.~\ref{fig:inpainting_comp}, previous inpainting models suffer from the problem of generating blurry content in the occluded areas. In addition, FuseFormer and E2FGVI also face serious image quality issues in non-occluded areas, making these methods difficult to apply in real-world video inpainting scenarios. On the other hand, our method maintains high consistency with warped videos in non-occluded areas while generating pleasing results in occluded areas.

\subsection{Ablation Study}
\begin{figure}
\centering
\includegraphics[width=1.0\linewidth]{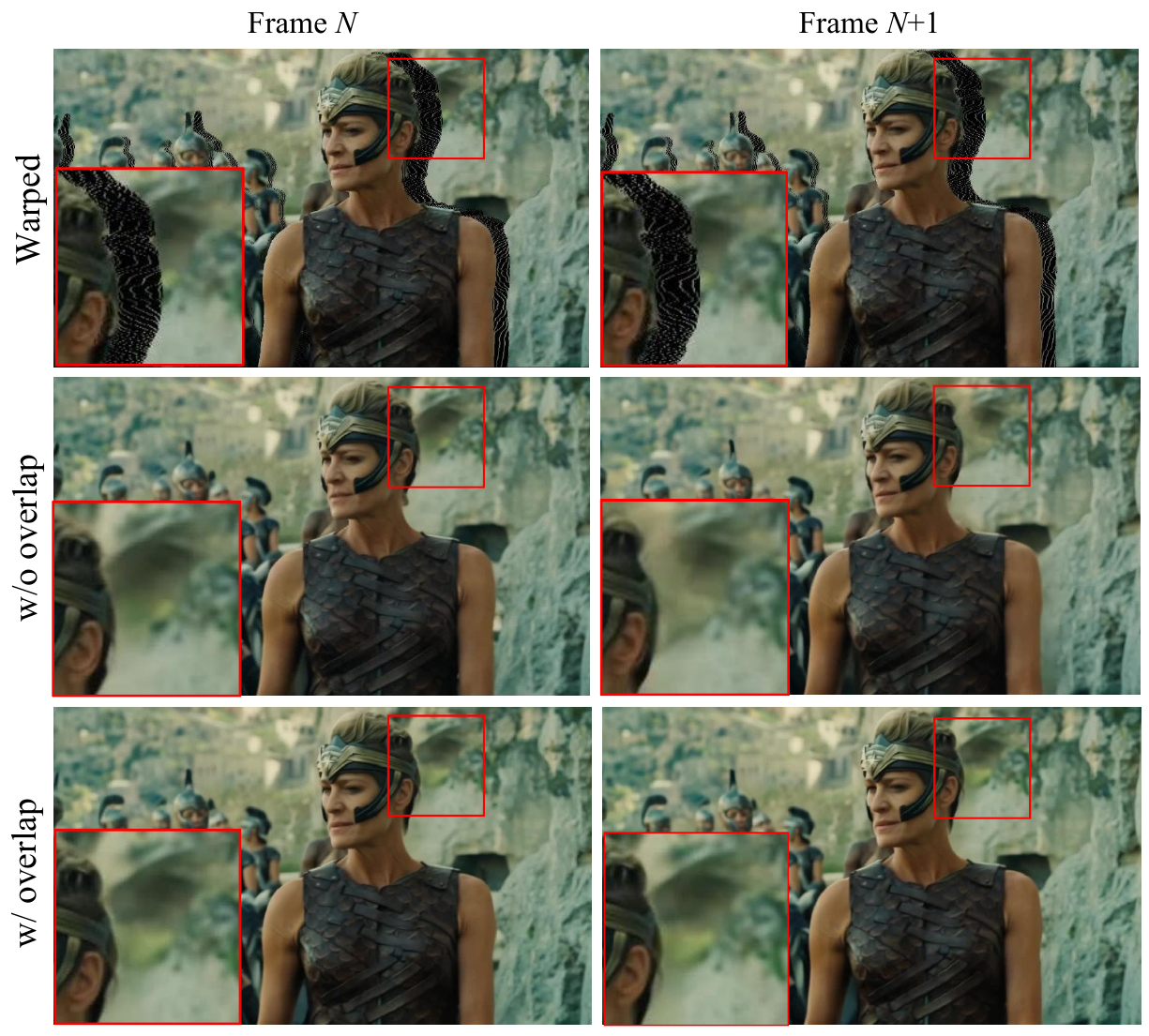}

\caption{Ablation results of auto-regressive modeling. We concatenate the last $n$ frames generated from the previous round with the warped frames of current round as input. When $n=0$, i.e., without overlap, the inpainting results of adjacent rounds cannot maintain temporal consistency, as shown in the second row.}
\label{fig:overlap_comparison}
\end{figure}

\textbf{Auto-regressive modeling.}  We evaluated the effectiveness of auto-regressive modeling through the following experiments: (1) \textbf{`w/o overlap'}, where each round of video frames is inferred independently and then concatenated together; (2) \textbf{`w/ overlap'}, where each round after the first uses the last $n$ frames of inpainting results from the previous round as input, with $n=3$ in this experiment. The experimental results are shown in Fig.~\ref{fig:overlap_comparison}. It can be observed that in the `w/o overlap' case, inconsistencies appear in the inpainting area between the front and back frames of adjacent rounds, while `w/ overlap' can solve this problem.

\begin{figure}
\centering
\includegraphics[width=1.0\linewidth]{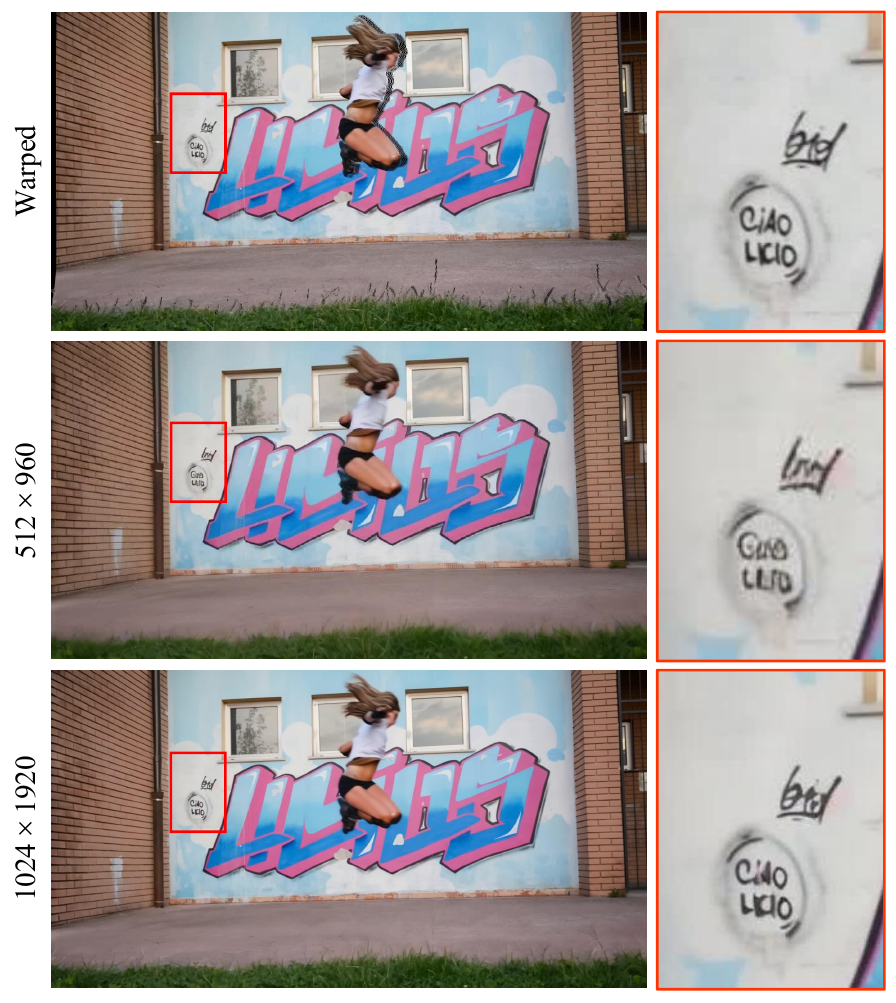}

\caption{Comparison results of our method at different resolutions. We use the tiled diffusion described in Sec.~\ref{subsec:inpainting} to handle high-resolution videos, Which maintains more details in the generated videos.}
\label{fig:tiling_comparison}
\end{figure}

\textbf{Tiled diffusion.} We validate the effectiveness of tiled diffusion through the following experiments: (1) when the resolution is low, e.g. $512\times 960$, perform global inpainting inference in the spatial dimension; (2) when the resolution is high, e.g. $1024\times 1920$, use tiled diffusion in the spatial dimension for multiple inferences and then obtain inpainting results through blending. The results are shown in Fig.~\ref{fig:tiling_comparison}. It can be observed that using tiled diffusion at high resolution can achieve better detailed results within the same memory constraints. Without tiled diffusion, inferring videos with a resolution of $1024\times 1920$ is not feasible due to the high memory usage of GPUs.

\section{Conclusion and Future Work}
\label{sub:conclusion}
%--------------------------------------
We have introduced a novel framework for converting 2D videos into stereoscopic 3D content to meet the growing demand for immersive digital experiences driven by advancements in VR and AR technologies. Our approach leverages foundation models as priors to enhance video depth estimation, achieving high-quality and detailed video depth maps. By combining video depth estimation, video splatting, and stereo video inpainting, our system successfully converts 2D videos into stereoscopic 3D videos that can be experienced with different devices like Apple Vision Pro.

\paragraph{Future Work.} While our proposed framework achieves promising results, several areas for future work remain to further enhance 2D-to-3D video conversion: (1) Future research could focus on developing more advanced depth estimation techniques that can provide even higher accuracy and consistency, particularly in challenging scenarios involving high motion or complex visual effects. (2) Optimization of the framework to support real-time video conversion would be a significant advancement, making the technology more practical for live streaming and real-time applications.

{
    \small
    \bibliographystyle{ieeenat_fullname}
    \bibliography{main}
}
% WARNING: do not forget to delete the supplementary pages from your submission
% \input{sec/X_suppl}

\end{document}